\documentclass{article}
\usepackage{spconf,amsmath,amssymb,graphicx,booktabs,multirow,xcolor}
\usepackage[hidelinks]{hyperref}
\newcommand{\best}[1]{\textbf{#1}}

\newcommand{\ns}{$^{\circ}$}
\newcommand{\worse}{$^{\dagger}$}

\title{Anchored Slice Propagation with Boundary Transport for 3D CT Segmentation from Three Orthogonal Annotated Slices}

\name{Tushar Kataria, Md Rakibul Haque, Shireen Y. Elhabian}
\address{Scientific Computing and Imaging Institute, University of Utah, Salt Lake City, USA \\ Kahlert School of Computing, University of Utah, Salt Lake City, USA \\
 { \tt\small \{tushar.kataria,rakibul,shireen\}@sci.utah.edu }
}

\begin{document}
\ninept
\maketitle

\begin{abstract}
Dense 3D CT annotation is costly, motivating methods that segment a volume from a few annotated slices. Sli2Vol propagates a binary mask through affinities between adjacent slices, but an error can persist because each prediction becomes the input to the next step. Three orthogonal annotations reduce dependence on any one chain, yet they offer more than three starting masks: where a chain crosses another annotated plane, the correct label is already known. Those labels could correct drift during propagation and indicate how much to trust each chain at fusion. Between annotated planes, the information carried by the chain may matter as well. A binary mask records foreground membership but discards distance to the boundary, a cue that could help follow gradual shape changes between slices.
We address these two sources of error with a fixed pretrained affinity backbone. We anchor each chain by correcting its prediction where it crosses another annotated plane, then use its agreement with those labels to weight fusion. We compare this anchored mask with two ways of transporting boundary geometry: a truncated signed-distance field (SDF) propagated by dense affinity averaging, and a contour whose vertices follow motion cues from the same affinity. The contour tracker starts from a hand-crafted rule and learns a motion correction from supervoxel pseudo-objects in 506 CT volumes, without manual organ labels. On 2,091 held-out structure instances from approximately 403 CT scans spanning 17 abdominal structures, anchoring raises mean Dice from 0.737 with orthogonal mask propagation to 0.769 and improves every structure. Beyond the anchored mask, learned contour tracking improves Dice by 5.7\% for the esophagus and 5.9\% for the postcava, while SDF transport improves right-adrenal Dice by 2.8\%. The anchored methods have similar mean Dice (0.769--0.772); contour tracking performs worse on the kidneys, stomach, and hepatic vessel tree. Existing annotations provide a consistent means of correcting propagation, while the benefit of boundary transport varies across anatomies. 
\end{abstract}


\section{Introduction}
\label{sec:intro}

Dense annotation of a 3D CT volume requires tracing a structure across many slices, especially when boundaries have poor contrast, cross-sections are small, or shape changes substantially. Slice-propagation methods~\cite{yeung2021sli2vol,bitarafan2022vol2flow,an2024sli2volplus,nihalaani2024estimation,haque2026inference} reduce this effort by transferring sparse annotations through correspondences between adjacent slices. Sli2Vol~\cite{yeung2021sli2vol}, for example, learns affinities without segmentation labels and propagates a binary mask from one annotated slice. Its predictions are sequential: a misplaced boundary or lost component can be carried into every subsequent slice.

Orthogonal annotations have been used for sparsely supervised 3D segmentation~\cite{cai2023orthogonal}. Our earlier slice-propagation work selected the largest cross-section along each axis, propagated three chains, and fused them with seed-distance weights~\cite{haque2026inference}. The three views reduce dependence on a single path, but each chain propagates independently before fusion. Every slice in one chain crosses the other two annotated planes along lines with known labels. A chain can therefore disagree with an available annotation and continue carrying the error. Cross-plane constraints have precedent in interactive 3D Livewire~\cite{poon2008efficient}. We call our use of these crossing labels \emph{anchoring}: they correct each propagated chain where the label is known and measure its agreement for weighted fusion. This makes fuller use of the same three annotations without retraining the affinity backbone.

Intersection constraints correct errors where labels are known, but each chain must still follow the anatomy between planes. In the slice-propagation methods considered here~\cite{yeung2021sli2vol,haque2026inference,an2024sli2volplus}, affinity averaging yields a soft prediction that is thresholded before the next step. If a boundary region is lost, the next step retains no record of its proximity to the object. This motivates transporting boundary geometry rather than foreground membership alone. Distance-based shape interpolation~\cite{raya1990shape} and signed-distance representations~\cite{osher2004level,park2019deepsdf,bogensperger2025flowsdf} suggest that boundary distance can capture gradual shape changes. We investigate whether propagating a truncated signed-distance field and reconstructing it after each step improves propagation between anchors. Alternatively, a contour follows the boundary as moving points rather than updating a field at every pixel. Drawing on video contour tracking~\cite{nam2021polygonal} and learned contour deformation~\cite{peng2020deep}, we estimate vertex motion from the same affinity while coupling neighboring vertices to limit irregular defo rmation. We correct the contour at annotated-plane crossings and train its motion correction on supervoxel pseudo-objects without manual organ labels, building on prior pseudo-label approaches~\cite{ouyang2022self,hansen2022anomaly}.

Sli2Vol+~\cite{an2024sli2volplus} learns correspondences from generated pseudo-labels, while SAM 2-based medical segmentation, MedSAM2, and nnInteractive use promptable models~\cite{dong2026segment,ma2025medsam2,isensee2025nninteractive}. These approaches leave open what known orthogonal intersections and alternative transported representations can add to the existing Sli2Vol affinity. We examine both with a fixed backbone and anchored-mask control. Validated on 342 cases, the methods are applied unchanged to 2,091 held-out structure instances from approximately 403 scans across 17 abdominal anatomies. Matched annotations distinguish anchoring gains from boundary-transport effects and reveal structures that remain difficult to propagate.
Our contributions are:
\begin{itemize}
\setlength{\itemsep}{0pt}
\item An anchor constraint and agreement-weighted fusion that use the three existing annotations throughout propagation.
\item Signed-distance and label-free contour transport compared with an anchored mask under a fixed affinity model.
\item A structure-level evaluation across 17 abdominal anatomies that separates consistent anchoring gains from anatomy-dependent boundary-transport effects. \href{To be Released on Acceptance}{ Code and models will be released via a public GitHub repository upon acceptance.}
\end{itemize}
\section{Method}
\label{sec:method}

Each volume has one annotated axial, coronal, and sagittal slice. Propagation in both directions along each axis produces three chains. For each chain, we use labels from the other two planes at their intersections to correct the propagated prediction. Agreement with those labels, measured before correction, also weights the chain during fusion. All methods use the same frozen affinity backbone, annotated slices, and intensity-verification procedure.

We compare three representations transported through this shared pipeline. The anchored-mask control propagates a binary indicator. The signed-distance field (SDF) variant averages truncated boundary distances through the affinity and reconstructs the field after each step. The contour variant estimates motion at boundary vertices, combines local displacements with shared shape motion, and re-extracts the contour for the next step. Comparing these variants isolates the effect of boundary transport from that of anchoring.

\textbf{Frozen affinity and orthogonal propagation.}
We use the pretrained Sli2Vol affinity backbone employed by orthogonal seeding~\cite{yeung2021sli2vol,haque2026inference}. For adjacent slices $k$ and $k+1$ along axis $a$, it produces an affinity $A_k^a(p,s)$ from each target pixel $p$ to source pixels $s$ in a local neighborhood, with $\sum_s A_k^a(p,s)=1$. Propagation of a scalar field $u_k^a$ is
\begin{equation}
\bigl(T_{A_k^a}u_k^a\bigr)(p)
=\sum_s A_k^a(p,s)u_k^a(s).
\label{eq:affinity_transport}
\end{equation}
The binary-mask baseline thresholds $T_{A_k^a}m_k^a$ to obtain the next mask. For each axis, the seed is the slice with the largest target cross-section, selected using the full ground-truth volume as in orthogonal seeding; realistic seed selection is not evaluated. The three chains use the original seed-distance weights $w_a(p)=\exp(-|k_a(p)-k_a^0|/\sigma)$, where $k_a^0$ is the seed location and $\sigma=30$ slices. All variants use the same backbone, seed locations, propagation directions, and Sli2Vol intensity verification. The backbone is never updated.

\textbf{Anchored propagation.}
For a slice in chain $a$, the other two annotated planes provide known labels along their lines of intersection. Let $\mathcal C_k^a$ contain the pixels on these lines and $y_k^a(p)$ their annotated labels. We constrain the propagated prediction on $\mathcal C_k^a$ before obtaining the next mask: mask transport takes the known foreground or background value, while SDF transport takes the corresponding negative or positive sign. The intersections also provide a measure of chain reliability. Let $\hat m_k^a$ be the raw, thresholded prediction before anchor correction and intensity verification. Its agreement with the known labels is
\begin{equation}
r_a(k)=\frac{1}{|\mathcal C_k^a|}
\sum_{p\in\mathcal C_k^a}
\mathbf{1}\!\left\{\hat m_k^a(p)=y_k^a(p)\right\}.
\label{eq:anchor_agreement}
\end{equation}
We smooth $r_a(k)$ with a five-slice moving average and floor the result at $0.05$; hereafter, $r_a(k)$ denotes this adjusted value. Let $q_a(p)\in[0,1]$ be the chain's foreground value supplied to fusion after anchor correction. Intensity verification instead acts on the hard mask used to start the next propagation step. Combining $r_a(k)$ with the seed-distance weight $w_a(p)$ gives
\begin{equation}
M(p)=\mathbf{1}\!\left\{
\frac{\sum_a w_a(p)\,r_a(k_a(p))\,q_a(p)}
{\sum_a w_a(p)\,r_a(k_a(p))}
>\frac{1}{2}
\right\}.
\label{eq:anchor_fusion}
\end{equation}
Thus, the existing annotations correct chains at their intersections and reduce the influence of chains that disagree with known labels. No additional annotations are required.

\textbf{Implicit boundary transport: signed-distance fields.}
Binary-mask transport retains only foreground membership. To transport boundary proximity as well, we convert the current mask to a truncated SDF $\phi_k^a$, negative inside the object:
\begin{equation}
\phi_k^a=
\operatorname{clip}\!\left(
d_{\mathrm{out}}(m_k^a)-d_{\mathrm{in}}(m_k^a),
-\tau,\tau
\right),
\qquad
\tilde\phi_{k+1}^a=T_{A_k^a}\phi_k^a.
\label{eq:sdf_transport}
\end{equation}
We constrain the sign of $\tilde\phi_{k+1}^a$ at annotated intersections and obtain the next mask from its zero level set. After the shared verification step, we recompute the truncated SDF from that mask before propagating again. Reinitialization prevents repeated affinity averaging from progressively smoothing the same field. For fusion, we convert the propagated SDF to a soft foreground prediction. We refer to this variant as \emph{SDF\,+\,anchors}, and to binary-mask transport with the same anchor mechanism as \emph{mask\,+\,anchors}.

\textbf{Explicit boundary transport: contour tracking.}
Inspired by object tracking across video frames, we also propagate the boundary directly. On slice $k$, the object is represented by a closed polygon $V_k^a=\{v_{k,i}^a\}_{i=1}^{N}$. The affinity is normalized over source pixels for each target pixel, so obtaining forward motion for a source vertex requires gathering target pixels that point back toward it. For vertex $v_{k,i}^a$, let $\mathcal Q(v_{k,i}^a)$ be its target search region. We form a forward distribution
\begin{equation}
B_{k,i}^a(p)=
\frac{A_k^a(p,v_{k,i}^a)}
{\sum_{q\in\mathcal Q(v_{k,i}^a)}
 A_k^a(q,v_{k,i}^a)},
\qquad p\in\mathcal Q(v_{k,i}^a),
\label{eq:forward_affinity}
\end{equation}
where affinity at a vertex is evaluated at its corresponding source pixel. Its expectation gives a candidate displacement,
\begin{equation}
\hat\delta_{k,i}^a=
\sum_{p\in\mathcal Q(v_{k,i}^a)}
B_{k,i}^a(p)\,p-v_{k,i}^a.
\label{eq:contour_displacement}
\end{equation}
The distribution's concentration supplies a confidence for the displacement estimate. We fit a confidence-weighted affine motion across vertices and add a smooth nonrigid residual, allowing deformation while discouraging irregular vertex motion. Annotated planes constrain the contour at known boundary crossings: nearby vertices move toward each crossing, with the correction decreasing along the contour. We then rasterize the corrected polygon. For fusion, we convert this rasterized prediction to a truncated SDF and apply the same sigmoid as in SDF transport, widening its transition as step confidence falls. Separately, intensity verification updates the hard rasterized mask; we extract the outer contours of all its components and resample their vertices at uniform arc length for the next step. These steps define the \emph{rule-based tracker}, which has no trained parameters. Its anchor-agreement score is measured on the polygon mask before crossing correction.

\textbf{Learned contour motion.}
We learn a correction to the rule-based motion rather than predicting vertex positions independently. For each vertex, a network $f_\theta$ receives its forward affinity distribution, candidate displacement and confidence, residual relative to the affine fit, and local contour geometry. It combines information across neighboring vertices and predicts a displacement correction. Additional outputs model uncertainty during training and indicate when the object has disappeared so that propagation can stop. The displacement output is initialized to zero, making the rule-based motion the starting point for learning.

\section{Results and Discussion}
\label{sec:exp}
\begin{table*}[!htb]
\centering\small\setlength{\tabcolsep}{1.0pt}
\caption{Development grid (342 cases): volumetric Dice, normalized surface Dice at 2\,mm (NSD; higher is better), and average symmetric surface distance (ASSD, mm; lower is better). Final columns report pooled paired differences from orthogonal seeding and the fraction of cases improved (parentheses); all reported differences have $p<10^{-9}$ against that reference. Sli2Vol uses one annotated slice; all other methods use the same three orthogonal slices. The italic rule-based tracker is the training-free initialization of the learned tracker.}
\label{tab:gate}
\begin{tabular}{@{}lccc@{\hskip 9pt}ccc@{\hskip 9pt}ccc@{\hskip 9pt}ccc@{}}
\toprule
 & \multicolumn{3}{c}{Liver, $n{=}20$} & \multicolumn{3}{c}{Pancreas, $n{=}281$} & \multicolumn{3}{c}{Spleen, $n{=}41$} & \multicolumn{3}{c}{vs orthogonal seeding, $n{=}342$} \\
\cmidrule(lr){2-4}\cmidrule(lr){5-7}\cmidrule(lr){8-10}\cmidrule(lr){11-13}
Method & Dice & NSD & ASSD & Dice & NSD & ASSD & Dice & NSD & ASSD & $\Delta$Dice (wins) & $\Delta$NSD (wins) & $\Delta$ASSD (wins) \\
\midrule
Sli2Vol, 1 seed~\cite{yeung2021sli2vol} & 0.909 & 0.680 & 5.90 & 0.530 & 0.408 & 8.18 & 0.884 & 0.794 & 3.38 & $-0.131$ (5\,\%) & $-0.118$ (6\,\%) & $+4.08$ (3\,\%) \\
Orthogonal seeds, mask~\cite{haque2026inference} & 0.950 & 0.775 & 2.36 & 0.681 & 0.539 & 3.69 & 0.922 & 0.838 & 1.81 & \multicolumn{3}{c}{reference} \\
Mask + anchors & 0.951 & 0.777 & 2.31 & 0.707 & 0.563 & 3.52 & 0.925 & 0.845 & 1.71 & $+0.022$ (\best{95\,\%}) & $+0.021$ (\best{93\,\%}) & $-0.16$ (83\,\%) \\
SDF + anchors & \best{0.954} & \best{0.794} & \best{1.96} & 0.723 & 0.583 & 3.10 & 0.922 & 0.811 & 1.62 & $+0.035$ (92\,\%) & $+0.034$ (89\,\%) & $-0.54$ (\best{88\,\%}) \\
\textit{Contour tracking, rule-based} & \textit{0.930} & \textit{0.671} & \textit{2.43} & \textit{0.716} & \textit{0.573} & \textit{\best{2.79}} & \textit{\best{0.937}} & \textit{\best{0.851}} & \textit{\best{0.93}} & \textit{$+0.029$ (70\,\%)} & \textit{$+0.024$ (63\,\%)} & \textit{\best{$-0.84$} (78\,\%)} \\
Contour tracking, learned & 0.941 & 0.714 & 2.19 & \best{0.745} & \best{0.603} & \best{2.79} & 0.929 & 0.820 & 1.10 & \best{$+0.053$} (84\,\%) & \best{$+0.047$} (77\,\%) & \best{$-0.84$} (81\,\%) \\
\bottomrule
\end{tabular}
\vspace{-1.5em}
\end{table*}
 
\textbf{Data and protocol.} We fixed all hyperparameter choices on 342 development cases: SLIVER07 liver (20)~\cite{heimann2009sliver}, Medical Segmentation Decathlon (MSD) Task07 Pancreas (281), and MSD spleen (41)~\cite{antonelli2022msd}. Held-out evaluation covered 17 structures and 2,091 structure instances from approximately 403 CT scans: 15 AMOS validation structures (93--100 instances each)~\cite{ji2022amos} and MSD hepatic tumor and vessel labels (303 each). We use the public Sli2Vol checkpoint from our orthogonal-seeding work~\cite{haque2026inference}, pre-trained without manual labels on 576 scans from CHAOS (40)~\cite{CHAOSdata2019}, MSD Liver test images (70)~\cite{simpson2019msd}, NIH Pancreas-CT (80)~\cite{roth2016data}, KiTS19 (210)~\cite{heller2019kits19}, and CT-LYMPH-NODES (176)~\cite{roth2014new}. Its weights remain fixed; contour training uses the 506 scans from the four sets excluding MSD Liver. Training and evaluation scan identifiers do not overlap, although patient identity across anonymized releases cannot be verified. All methods except single-seed Sli2Vol share three annotated slices. We report native-resolution Dice, plus 2\,mm normalized surface Dice (NSD) and average symmetric surface distance (ASSD, mm) on the development set. Paired two-sided Wilcoxon tests use orthogonal seeding and, where stated, the anchored mask as references.

\textbf{Main results.} Anchoring and boundary transport yield their clearest development-set gains on the pancreas, where orthogonal mask propagation achieves 0.681 Dice (Table~\ref{tab:gate}). Anchoring its mask raises Dice to 0.707, anchored SDF transport to 0.723, and learned contour tracking to 0.745. Relative to orthogonal seeding, the two boundary methods improve pancreatic Dice by 6.2\% and 9.4\%, respectively. The gains extend to surface accuracy: NSD rises from 0.539 to 0.583 with SDF transport and 0.603 with learned contour tracking, while ASSD falls from 3.69 to 3.10 and 2.79\,mm. Together, these results show that anchoring and boundary transport improve both overlap and boundary accuracy on the pancreas, where orthogonal mask propagation remains less accurate.

Across the 342-case development grid, anchored SDF and learned contour tracking improve Dice, NSD, and ASSD over orthogonal seeding (paired $p<10^{-9}$ for each metric). Against the matched mask + anchors control (paired Wilcoxon, $p<0.05$), SDF improves all three metrics on liver and pancreas but worsens spleen NSD. Learned contour tracking improves all three metrics on pancreas and the pooled grid, and improves spleen ASSD, but worsens liver Dice and NSD. Rule-based tracking improves spleen Dice, pancreas NSD, and ASSD on pancreas, spleen, and the pooled grid, while worsening liver Dice and NSD. The learned tracker has the best pooled Dice and NSD; its Dice win rate against orthogonal seeding is 84\%, compared with 70\% for the rule-based tracker. Liver and spleen already have high Dice with orthogonal seeding and show smaller or less consistent gains. On this development grid, the largest improvements occur on the pancreas.

Table~\ref{tab:chains} shows whether gains arise within individual chains or only after fusion. The axial chain of orthogonal seeding matches single-seed Sli2Vol (Dice 0.595); fusing three unanchored chains raises Dice to 0.725, a gain of 0.130. Anchoring improves all three chains and raises fused Dice to 0.748, suggesting that correcting predictions during propagation contributes beyond combining independent views. Relative to this matched control, SDF transport improves every chain and raises fused Dice to 0.760; its gain is therefore also present before fusion. The rule-based contour tracker improves axial Dice from 0.627 to 0.652 but reduces coronal Dice from 0.588 to 0.543, yielding 0.755 after fusion. This directional difference suggests that a fixed vertex-motion rule does not read the affinity equally well along every axis. Learning improves the contour chains to 0.657 axially, 0.587 coronally, and 0.630 sagittally, from 0.652, 0.543, and 0.593 for the rule-based tracker. The larger coronal and sagittal gains indicate where the learned motion correction helps most; fused Dice reaches 0.778 on the development grid.

\begin{table}[!htb]
\vspace{-1em}
\centering\small\setlength{\tabcolsep}{4pt}
\caption{Chain ablation on 342 development cases. Dice is reported for each axis before fusion and for the fused prediction. Anchoring improves all three mask-propagation chains relative to orthogonal seeding; boundary transport changes their performance differently across axes.}
\label{tab:chains}
\begin{tabular}{@{}lcccc@{}}
\toprule
Method & axial & coronal & sagittal & fused \\
\midrule
Sli2Vol, 1 seed & 0.595 & -- & -- & 0.595 \\
Orthogonal seeds, mask & 0.595 & 0.514 & 0.527 & 0.725 \\
Mask + anchors & 0.627 & 0.588 & 0.570 & 0.748 \\
SDF + anchors & 0.637 & \best{0.596} & 0.601 & 0.760 \\
\textit{Contour tracking, rule-based} & \textit{0.652} & \textit{0.543} & \textit{0.593} & \textit{0.755} \\
Contour tracking, learned & \best{0.657} & 0.587 & \best{0.630} & \best{0.778} \\
\bottomrule
\end{tabular}
\vspace{-1em}
\end{table}

\begin{table*}[!htb]
\centering\scriptsize\setlength{\tabcolsep}{1.5pt}
\caption{Held-out volumetric Dice for 17 structures (2,091 structure instances from approximately 403 CT scans). All methods share a frozen affinity; all except single-seed Sli2Vol use three orthogonal annotations. Anchors are applied only where indicated. Bold marks the best value per structure; Mean is unweighted across structures. Pro: prostate/uterus; GB: gall bladder; PCv: postcava; LAd/RAd: adrenal glands; HTu/HVe: hepatic tumor/vessel. Relative to orthogonal seeding, \ns{} indicates no significant difference and \worse{} a significant loss; unmarked comparison values indicate improvement (paired Wilcoxon, $p<0.05$).}
\label{tab:cohort}
\begin{tabular}{@{}lcccccccccccccccccc@{}}
\toprule
Method & Liv & Spl & LKid & RKid & Sto & Bla & Pro & Pan & GB & Duo & Eso & Aor & PCv & LAd & RAd & HTu & HVe & Mean \\
$n$ & 100 & 100 & 100 & 98 & 99 & 99 & 97 & 100 & 93 & 99 & 100 & 100 & 100 & 100 & 100 & 303 & 303 & 17 \\
\midrule
Sli2Vol, 1 seed~\cite{yeung2021sli2vol} & 0.857\worse & 0.878\worse & 0.882\worse & 0.890\worse & 0.572\worse & 0.671\worse & 0.648\worse & 0.587\worse & 0.608\worse & 0.453\worse & 0.415\worse & 0.679\worse & 0.442\worse & 0.598\worse & 0.509\worse & 0.619\worse & 0.474\ns & 0.634 \\
Orthogonal seeds, mask~\cite{haque2026inference} & 0.894 & 0.921 & 0.910 & 0.914 & 0.707 & 0.739 & 0.764 & 0.729 & 0.718 & 0.577 & 0.547 & 0.825 & 0.722 & 0.724 & 0.641 & 0.718 & 0.487 & 0.737 \\
Mask + anchors & 0.897 & 0.924 & \best{0.914} & \best{0.917} & \best{0.726} & 0.768 & 0.815 & 0.743 & 0.765 & 0.615 & 0.616 & 0.838 & 0.741 & 0.777 & 0.724 & 0.765 & \best{0.536} & 0.769 \\
SDF + anchors & 0.896 & 0.920\ns & 0.911\ns & 0.917 & 0.710\ns & 0.780 & 0.817 & 0.740 & 0.778 & 0.613 & 0.613 & \best{0.843} & 0.761 & 0.775 & 0.744 & 0.776 & 0.499 & 0.770 \\
\textit{Contour tracking, rule-based} & \textit{\best{0.916}} & \textit{\best{0.931}} & \textit{0.892\worse} & \textit{0.892\worse} & \textit{0.695\worse} & \textit{\best{0.806}} & \textit{0.829} & \textit{\best{0.747}} & \textit{0.787} & \textit{0.607} & \textit{0.626} & \textit{0.816\ns} & \textit{0.777} & \textit{\best{0.789}} & \textit{\best{0.749}} & \textit{0.795} & \textit{0.410\worse} & \textit{0.769} \\
Contour tracking, learned & 0.905 & 0.927 & 0.894\worse & 0.897\worse & 0.701\ns & 0.801 & \best{0.842} & 0.745 & \best{0.799} & \best{0.617} & \best{0.651} & 0.840 & \best{0.785} & 0.787 & 0.725 & \best{0.798} & 0.410\worse & \best{0.772} \\
\bottomrule
\end{tabular}
\end{table*}
\begin{figure*}[!htb]
    \centering
    \includegraphics[width=1.0\linewidth]{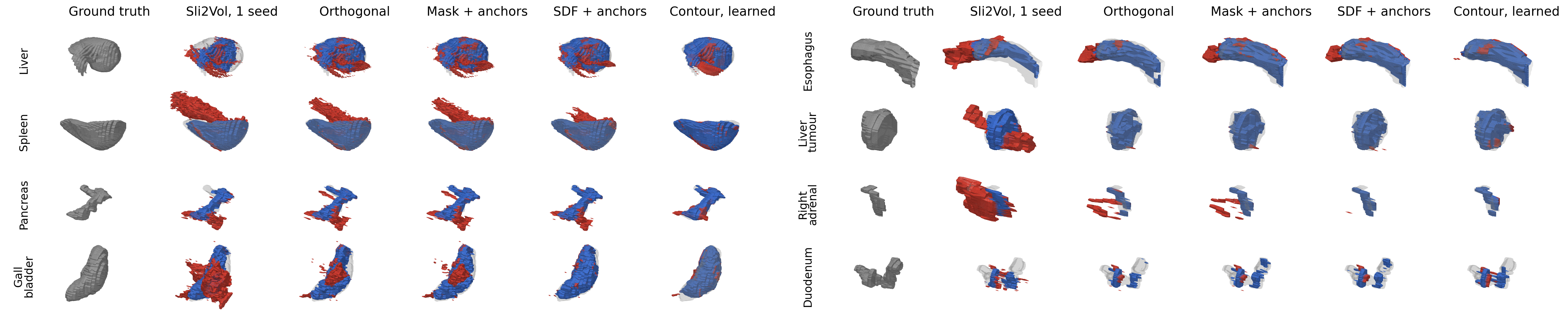}
\caption{Held-out 3D renderings, one case per row. All methods share a frozen affinity; all but Sli2Vol use three annotated slices. Gray denotes ground truth; predictions are blue within it and red outside it. Uncovered gray indicates missed anatomy. Left: liver, spleen, pancreas, and gall bladder. Right: esophagus, hepatic tumor, right adrenal gland, and duodenum, where all methods lose the structure before the loop.}
    \label{fig:qual}
    \vspace{-1.5em}
\end{figure*}
\textbf{Held-out cohort evaluation.} Table~\ref{tab:cohort} evaluates every method unchanged on 2,091 structure instances spanning 17 structures. Gains are largest where the baselines remain inaccurate. Esophagus Dice rises from 0.415 with Sli2Vol to 0.547 with orthogonal seeding, 0.616 with an anchored mask, and 0.651 with learned contour tracking; postcava Dice rises from 0.442 to 0.722, 0.741, and 0.785. For the right adrenal gland, anchored SDF transport reaches 0.744, versus 0.641 with orthogonal seeding and 0.724 with an anchored mask. Boundary transport therefore adds to anchoring on structures that remain difficult after three-axis mask fusion.

Overall, anchoring raises mean Dice from 0.737 to 0.769 and improves every structure. Anchored SDF transport reaches 0.770, and learned contour tracking reaches 0.772; these unweighted means are close. Against mask + anchors, paired Wilcoxon tests ($p<0.05$) show significant Dice gains/losses on 6/4 structures for SDF transport, 9/5 for rule-based contour tracking, and 8/4 for learned contour tracking. The learned tracker gains roughly three to four Dice points over the anchored mask on the bladder, esophagus, gall bladder, hepatic tumor, and postcava. The anchored mask remains stronger on the kidneys and stomach and performs best on the hepatic vessel tree. When structure instances are weighted equally, the anchored mask also scores higher than the learned tracker (0.746 versus 0.739). Figure~\ref{fig:qual} shows representative results and a duodenum failure.

Important failures remain despite the overall gains. The best Dice achieved by any method is 0.536 for the hepatic vessel tree, 0.617 for the duodenum, and 0.651 for the esophagus. Although the contour tracker re-extracts multiple components, its vessel Dice is 0.410, below the anchored mask's 0.536; maintaining thin branches across propagation steps remains difficult. The duodenum and esophagus also have narrow or changing cross-sections that are difficult to follow using adjacent-slice affinity alone. Future work could combine propagation with volumetric alignment, using predictions where chains agree, and investigate cross-slice connectivity constraints for vessels.

\textbf{Qualitative results.} In Fig.~\ref{fig:qual}, single-seed Sli2Vol loses the boundary and leaks on the spleen, gall bladder and adrenal gland. Orthogonal seeding removes most leakage but leaves fragments where chains disagree. Anchors remove fragments crossing annotated planes; SDF and contour transport further reject leaked tissue on the gall bladder, esophagus and adrenal gland. The pancreatic tail leak also shrinks across the row. Predictions are nearly indistinguishable on the liver, where orthogonal seeding is already accurate (Table~\ref{tab:cohort}). On the duodenum, every method stops before the loop, leaving ground truth uncovered.

\textit{Limitations.} We did not evaluate a soft-mask transport control or compare with Sli2Vol+~\cite{an2024sli2volplus} and SAM 2-style methods under the same three-slice budget. Seed positions were selected from ground truth rather than through a realistic annotation workflow. Although the tracker re-extracts multiple contour components, its poor hepatic-vessel result shows that it does not reliably preserve branching structures during propagation. Future work will evaluate soft-mask transport and matched three-slice baselines, study practical seed selection, and improve tracking of thin vessel branches.

\vspace{-1em}
\section{Conclusion}
This study shows that three annotated slices can guide propagation beyond its starting points. Using their intersections to correct chains and weight fusion improves Dice over orthogonal seeding on all 17 held-out structures with the affinity backbone fixed. Transporting boundary geometry adds gains that depend on anatomy: learned contour tracking helps the esophagus and postcava, while signed-distance transport helps the right adrenal gland. Neither improves the cohort uniformly, and the anchored mask remains strongest on the hepatic vessel tree. The consistent benefit comes from using known labels throughout propagation; the choice of what to transport matters most at the level of individual structures.

\textbf{Compliance with Ethical Standards.} This retrospective study used publicly available CT data from CHAOS~\cite{CHAOSdata2019}, NIH Pancreas-CT~\cite{roth2016data}, KiTS19~\cite{heller2019kits19}, CT-LYMPH-NODES~\cite{roth2014new}, SLIVER07~\cite{heimann2009sliver}, the Medical Segmentation Decathlon~\cite{antonelli2022msd}, and AMOS~\cite{ji2022amos}. Ethics approval was not required for this secondary analysis.

\textbf{Acknowledgments.} The authors thank the Kahlert School of Computing and the Scientific Computing and Imaging Institute at the University of Utah for their support. This work received no external funding. ChatGPT and Claude Code assisted with manuscript editing, inference-script development, checking reported values against experimental outputs, and formatting tables in LaTeX. The authors reviewed the manuscript, code, and analyses and take responsibility for the reported findings. The authors declare no financial or non-financial conflicts of interest.
{
\bibliographystyle{IEEEbib}
\bibliography{refs}
}
\end{document}